\documentclass[twocolumn,10pt]{article}

\usepackage{graphicx}
\usepackage{siunitx}

\usepackage{hyperref}   
\hypersetup{
	colorlinks=true,
	linkcolor=blue,
	citecolor=blue,
	urlcolor=blue,
}
\usepackage{mathtools}
\usepackage{amsmath}
\usepackage{lipsum}
\usepackage{diagbox}

\usepackage{tabularx}
\usepackage{array}
\usepackage{epsfig} 
\usepackage{lipsum}
\usepackage{xspace}
\usepackage{amsmath}
\usepackage{booktabs}
\usepackage{rotating}

\usepackage[style=apa,sortcites=true,sorting=nyt,backend=biber,hyperref=true]{biblatex}
\usepackage{authblk}

\usepackage{mathptmx}

\usepackage{caption}
\usepackage{geometry}
\let\oldparagraph\paragraph

\renewcommand{\paragraph}[1]{\oldparagraph{#1.}}

\newcommand{\Loss}{\mathcal{L}} 
\newcommand{\MATLAB}{\textsc{Matlab}\xspace} 

\title{Deep Generative Model-based Synthesis of \\Four-bar Linkage Mechanisms with Target Conditions}
\date{}

\author[1]{Sumin Lee$^*$}
\author[1]{Jihoon Kim$^*$}
\author[2,3]{Namwoo Kang$^\dagger$}

\affil[1]{Department of Mechanical Engineering, KAIST}
\affil[2]{Cho Chun Shik Graduate School of Mobility, KAIST}
\affil[3]{Narnia Labs}

\begin{document}

\maketitle

\begingroup
\renewcommand{\thefootnote}{$^*$} 
\footnotetext{Equal contribution.}
\renewcommand{\thefootnote}{$^\dagger$} 
\footnotetext{Corresponding author: nwkang@kaist.ac.kr}
\endgroup

\begin{abstract}
Mechanisms are essential components designed to perform specific tasks in various mechanical systems. However, designing a mechanism that satisfies certain kinematic or quasi-static requirements is a challenging task. The kinematic requirements may include the workspace of a mechanism, while the quasi-static requirements of a mechanism may include its torque transmission, which refers to the ability of the mechanism to transfer power and torque effectively. In this paper, we propose a deep learning-based generative model for generating multiple crank-rocker four-bar linkage mechanisms that satisfy both the kinematic and quasi-static requirements aforementioned. The proposed model is based on a conditional generative adversarial network (cGAN) with modifications for mechanism synthesis, which is trained to learn the relationship between the requirements of a mechanism with respect to linkage lengths. The results demonstrate that the proposed model successfully generates multiple distinct mechanisms that satisfy specific kinematic and quasi-static requirements. To evaluate the novelty of our approach, we provide a comparison of the samples synthesized by the proposed cGAN, traditional cVAE and NSGA-II. Our approach has several advantages over traditional design methods. It enables designers to efficiently generate multiple diverse and feasible design candidates while exploring a large design space. Also, the proposed model considers both the kinematic and quasi-static requirements, which can lead to more efficient and effective mechanisms for real-world use, making it a promising tool for linkage mechanism design.
\end{abstract}



\setcounter{section}{0}
\section{Introduction}
Mechanism design is essential in many mechanical systems to perform tasks. 
For example, many different forms of a mechanism are used in locking pliers, suspensions, robots, and more.
In particular, a mechanism with linkages is widely used in industries due to its ability to transfer a simple circular motion (e.g., from a motor) into multi-joint movements that draw a unique path or a workspace.
Therefore, a linkage mechanism can be very useful if its design is carefully considered and tested for one's requirements.
Nevertheless, it is significantly challenging to design a multi-linkage mechanism that satisfies specific requirements, since its behavior changes significantly with minor changes in its linkage lengths \parencite{tsai2004kinematic, lee1999generalized}.
Thus, the mechanism synthesis is a very complex problem and has not been well understood by the field. 
The synthesis has often been done simply by trial and error, especially in industries, which makes it extremely difficult to find an optimal mechanism that suits one’s requirements.
Although it is possible to analytically calculate the lengths of a linkage mechanism that passes a few given points, e.g., Burmester Theory~\parencite{burmester1888theory,deshpande2017burmester, shin2023optimal}, real-world problems and requirements are often more complicated.

There has been significant effort in solving this issue in the field.
The numerical methods have been studied to find the lengths of linkages that follow a specific target path with feed-forward optimization processes ~\parencite{kang2016topology, yim2019linkageTO, yu2020linkageTOspring, han2021linkageTO}.
These methods are effective in providing a good solution if there is a linkage mechanism that can definitely satisfy these conditions.
However, their iterative processes can be time-consuming, especially in high resolution, which is more problematic when these processes must be repeated for different given conditions.
Further, these methods only can provide a single solution to the problem where there could be many other viable linkage mechanisms with similar conditions.

The data-driven methods are also an active research area~\parencite{yim2021big}.
\parencite{heyrani2022links} created a big dataset of planar linkage mechanisms for data-driven kinematic design.
Their data includes the paths drawn by the mechanisms, which provides a unique opportunity to simply find mechanisms from the dataset that satisfy required kinematic conditions.
Nevertheless, they mentioned that the generation of the dataset is very computationally expensive, and there still may not be a mechanism that perfectly satisfies the target path.
The deep learning-based~\parencite{deshpande2019VAE} and also image-based~\parencite{deshpande2021image} generative methods have been studied to synthesize paths of multiple linkage mechanisms.
These methods are useful in that they are able to generate samples more quickly than traditional optimization methods discussed above.
Also, they are able to generate a wide range of solutions, whereas most of the optimization-based methods can only generate a single sample.
However, the quality of the generated samples may be inferior compared to the optimization-based methods as it relies on the limited amount of the dataset it was trained on.

While reviewing the past work, we realized that most of the studies in four-bar linkage mechanism synthesis focus on generating a mechanism that can follow a specific target path~\parencite{han2021linkageTO, heyrani2022links, deshpande2019VAE, deshpande2019auto}.
However, many real-world problems do not need the whole path of a mechanism to be considered~\parencite{raghavan2004suspension, wongratanaphisan2007suspension,  wang2014gripper}, and it may be more useful to have multiple mechanisms that satisfy various requirements that are required to consider for real-world use cases than only strictly considering their paths. 
For example, mechanism synthesis models that only consider kinematic conditions, such as paths, may create mechanisms that require very high torque to perform tasks, which is infeasible to use in the real world.
Thus, it is important to design a mechanism that considers kinematic and quasi-static conditions simultaneously.
All the synthesis methods mentioned above do not consider any quasi-static or dynamic conditions in linkage mechanisms (e.g., torque transmission, maximum payload, etc.) that are essential to consider for a mechanism to perform meaningful tasks in the real application.
Therefore, implementing the application requires additional tweaks and compromises in design performance.
To effectively overcome these problems, we propose a methodology for condition-based linkage mechanism synthesis. The proposed methodology is based on a conditional generative adversarial network (cGAN), a deep learning-based generative model capable of synthesizing multiple diverse linkage mechanism samples that satisfy the given kinematic and quasi-static requirements. 
A traditional approach for mechanism synthesis solely considers the whole path of a mechanism with an optimization-based algorithm. 
However, we argue that this approach hugely hinders the diversity of the optimized mechanisms, considering we often do not need to optimize the whole path of a mechanism.
Thus, in order to increase the diversity of generated linkage mechanisms, we have formulated simplified kinematic and quasi-static conditions where there could be multiple viable linkages even when the same conditions are given. This allows for the generation of mechanisms with various design and paths that satisfy the same target conditions. 
Designers then will be able to choose a mechanism that suits their needs from multiple samples that satisfy the requirements. 

\begin{figure*}[t]
    \centering
    \includegraphics[width=\textwidth]{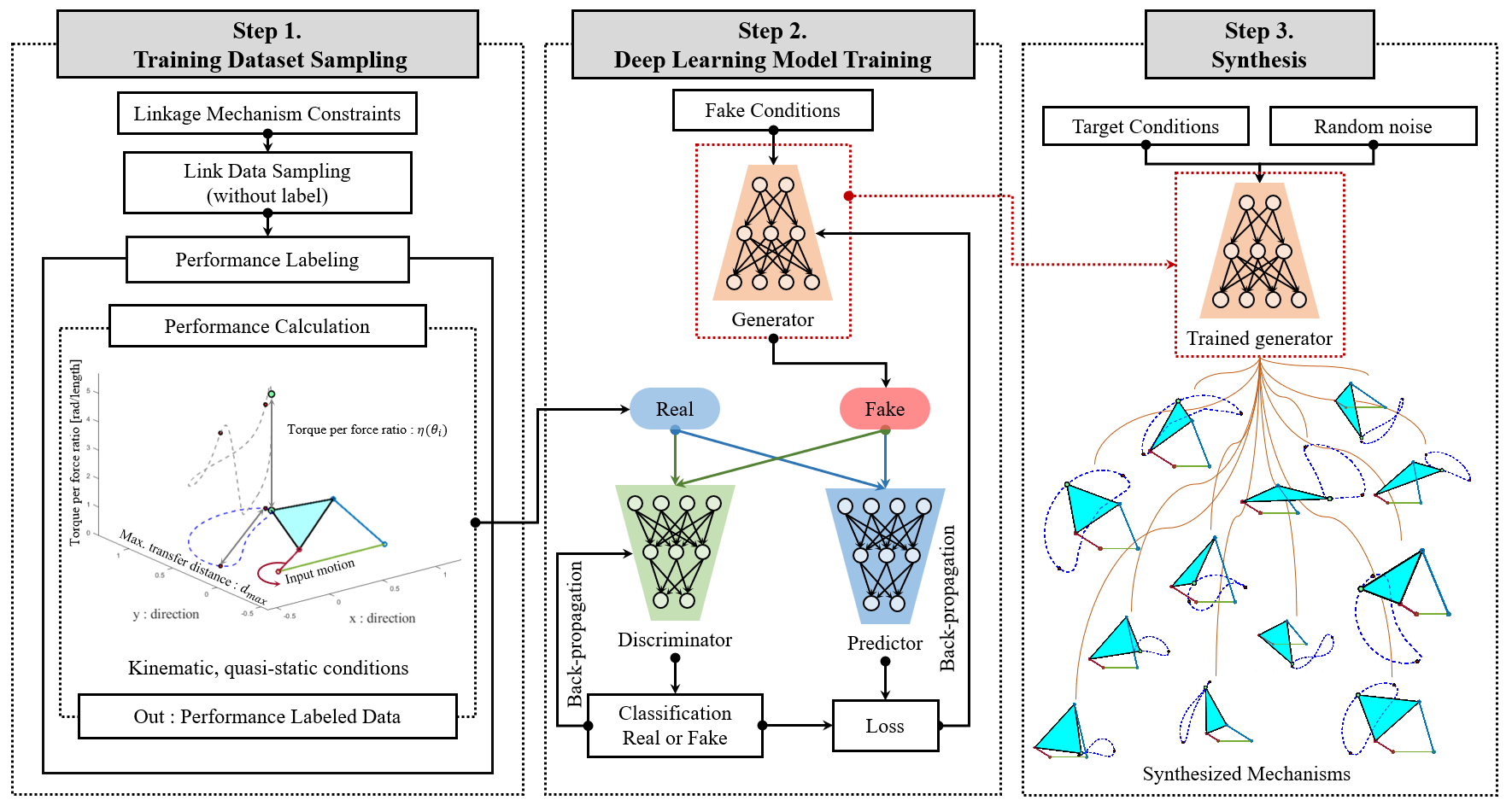}
    \caption{ 
    Deep generative model-based mechanism synthesis considering both kinematic and quasi-static conditions
    }
    \label{fig:overall-flow}
\end{figure*}

Figure~\ref{fig:overall-flow} shows the overall flow for the training of the proposed generative model and the linkage mechanism synthesis.
This paper is structured as follows:
Section~\ref{sec:related-work} provides a brief overview of the previous work on mechanism synthesis and deep generative models.
In Section~\ref{sec:generation-of-linkage-mechanisms}, the mechanism samples are generated for the training of the deep generative model.
In Section~\ref{subsec:structure-and-training-of-cGAN}, the generative model is trained with this dataset and its synthesis performance is evaluated in Section~\ref{subsec:cGAN-evaluation}.
An ablation study on cGAN was performed in Section~\ref{sec:ablation-studies} to evaluate the effectiveness of the modifications made on the generative model to ensure good mechanism synthesis.
The contents of this paper is summarized in Section~\ref{sec:conclusion}.

\section{Related Work}\label{sec:related-work}
The field of mechanism synthesis has been matured with many different methodologies proposed over the past few decades, ranging from early analytical methods \parencite{burmester1888theory,sandor1959analytical,erdman1981precisionpoint} to the latest deep learning-based generative approaches.
Here, we briefly cover the past work that has influenced this paper.

\subsection{Numerical Methods}
There has been significant effort in the early studies to use numerical optimization algorithms to perform mechanism synthesis.
\cite{cabrera2002GA} used genetic algorithm (GA), an optimization algorithm inspired by the process of natural selection and evolution, to perform optimal synthesis of planar mechanisms.
\cite{acharyya2009EAcomparison} further performed a performance comparison of three different evolutionary algorithms: GA, particle swarm optimization (PSO), and differential evolution (DE) for four-bar mechanism synthesis.

In terms of representing motion of mechanisms, \parencite{kim2007springblock} proposed a unified planar linkage model consisting of rigid blocks with springs for optimal mechanism synthesis \parencite{han2021linkageTO}.
Further, \cite{ullah1997fourier} used Fourier descriptors to form an effective objective function that is used for optimal mechanism synthesis with a stochastic global search method.
\cite{li2022fourier} also used Fourier descriptors to represent planar motion of mechanisms.

\subsection{Deep Learning-based Methods}
Thanks to the recent rise in deep learning-based generative models (DGMs), many studies have attempted to use DGMs for mechanism synthesis.
The authors refer to \parencite{regenwetter2022review} for a detailed review of DGMs in engineering design.
Below covers the studies on DGMs for engineering design and the applications on mechanism synthesis. 

\paragraph{Variational Autoencoders (VAE)}
VAEs \parencite{kingma2013vAE} are a variant of autoencoders, which are neural network architectures commonly used for dimensionality reduction, feature learning, and data compression. 
VAEs can generate new samples that are similar to the ones they were trained on, using its encoder and decoder network. 
\cite{deshpande2019VAE} used VAE to learn the probability distribution of possible linkage parameters to relate user inputs with corresponding linkage parameters, and use them for kinematic mechanism synthesis.
Their work was further improved using a convolutional VAE that enables an image-based variational path synthesis \parencite{deshpande2020convVAE}.

\paragraph{Generative Adversarial Networks (GAN)}
GANs \parencite{goodfellow2020gan} consist of two neural networks, the generator and the discriminator, in an adversarial relationship while training.
The generator takes random noise or a latent vector as input and generates samples that resemble the real data, whereas the discriminator acts as a binary classifier between real and generated samples.
They are iteratively and adversarily trained until the discriminator is unable to tell the difference between the two.
Once the GAN is trained, you can sample from the generator by feeding random noise or latent vectors, which produces samples that are similar to the real data.
Both VAEs and GANs are very popular DGMs that are increasingly used for engineering design.
However, they are different in that VAEs are probabilistic models that explicitly model the underlying probability distribution of data and focus on representation learning, while GANs are adversarial models with an implicit latent space that aim to generate highly realistic data. 

Due to its powerful synthesis performance in the field of computer vision \parencite{wang2021GANsurvey}, many studies worked on using GANs for engineering design.
Topology optimization, a method used the best material distribution to achieve desired performance objectives while minimizing constraints, traditionally relied on iterative numerical optimization algorithms.
Many studies have attempted to improve this limitation using machine and deep learning techniques \parencite{shin2023TOreview}, especially using GAN for improved diversity, both for 2D \parencite{oh2019deep,yu2019deep} and 3D \parencite{nie2021topologygan}.

Further, numerous studies have focused on modifying GANs for engineering design.
PaDGAN \parencite{chen2021padgan} introduced a modified loss function that motivates diversity, quality and design domain coverage of generated designs with determinantal point processes (DPP) \parencite{kulesza2012DPP} and performance estimators.
PcDGAN \parencite{heyrani2021pcdgan} is similar to PaDGAN in that DPP is used to improve sample diversity, and is also similar to continuous conditional GAN (CcGAN) \parencite{ding2020ccgan} for synthesis with continuous input variables given as conditions.

Albeit these improvements and unique advantages of cGAN, as far as we are concerned, there have been no attempts to use cGAN for mechanism synthesis.
Thus, this study also focuses on the viability of using a cGAN model for mechanism synthesis.
Because the aim of this study is to assess the viability of a cGAN model for the synthesis of mechanisms considering multiple conditions, the baseline cGAN model was used with minor modifications, rather than testing and picking a specific state-of-the-art GAN variant.


\section{Generation of Linkage Mechanisms}
\label{sec:generation-of-linkage-mechanisms}

\begin{figure}[tb]
    \centering
    \includegraphics[width=0.45\textwidth]{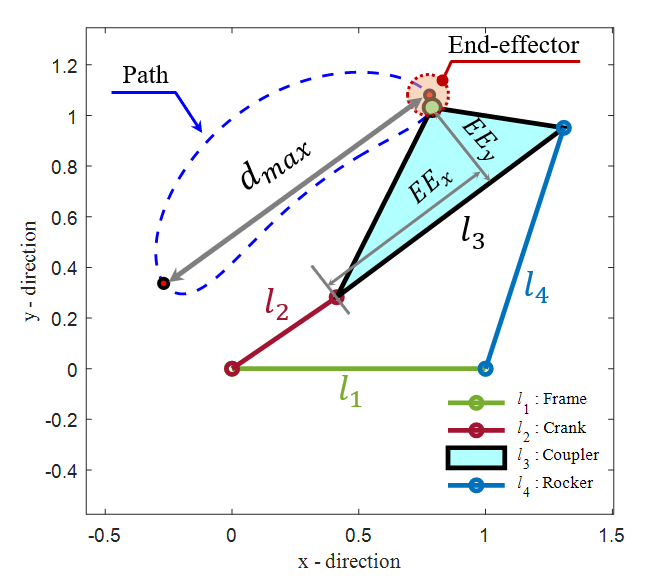}
    \caption{A labeled example of a crank-rocker four-bar linkage mechanism}
    \label{fig:linkage-workspace}
\end{figure}

\subsection{Crank-rocker Mechanism}
\label{subsec:linkage-mechanism}
It is crucial that the generated samples satisfy the constraints of a four-bar linkage mechanism.
While there are many types of four-bar linkage mechanisms (e.g., drag-link, crank-rocker, double-rocker, etc.), this paper only considers a crank-rocker mechanism as its crank can fully revolve and continuously draw a path as shown in Fig.~\ref{fig:linkage-workspace}. 
This provides the data needed to calculate the kinematic and quasi-static conditions for the inverse design of mechanisms.


\begin{figure}[tb]
    \centering
    \includegraphics[width=0.45\textwidth]{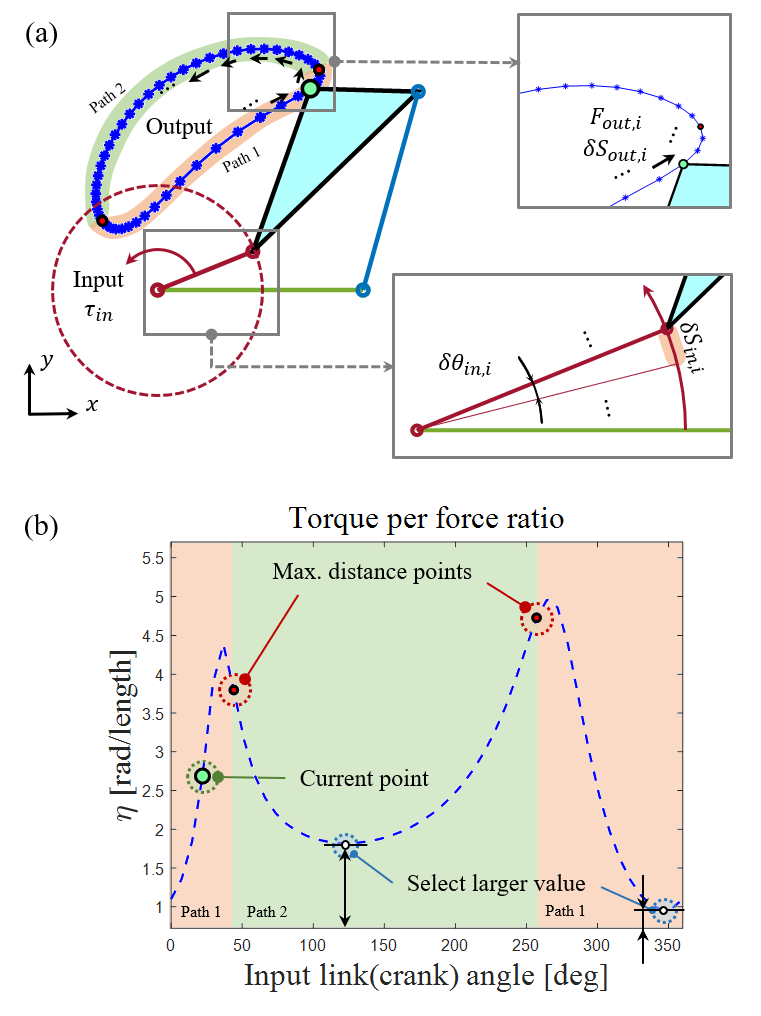}
    \caption{(a) The parameters needed for the calculation of $\eta_{min}$ and (b) the plot of $\eta$ along the path of a mechanism}
    \label{fig:vis-eta-min}
\end{figure}

According to Grashof's Law, in order for a crank-rocker system to rotate continuously, the sum of the shortest and longest link lengths is less than or equal to the sum of the other two link lengths.
Applying Reuleaux's Law, it must satisfy all the conditions in Eq.~(\ref{eq:linkage-conds}).
Since $l_1$ is fixed to $1.0$, the other linkage lengths can be considered as ratios with respect to $l_1$.

\begin{equation}
    \label{eq:linkage-conds}
    \begin{aligned}
l_3 + l_2 < l_1 + l_4  
    \end{aligned}
\end{equation}

In theory, any fully-rotating crank-rocker mechanisms can be produced if they satisfy the conditions in Eq.~(\ref{eq:linkage-conds}).
However, it is not viable to create a mechanism too long that would be infeasible or even impossible to produce in the real world, as there are infinite number of link sets that can satisfy Eq.~(\ref{eq:linkage-conds}). 
Thus, the range of of the linkage lengths of crank-rocker mechanisms were arbitrarily selected as shown in Table~\ref{tab:length-range} to have an appropriate range based on the experience. It is important to note that this range can be freely adjusted to suit different applications.

\begin{table}[htb]
\caption{The ranges of linkage lengths set to generate mechanisms in meters}
\label{tab:length-range}
\begin{center}
\begin{tabularx}{\linewidth}{c|*{5}{>{\centering\arraybackslash}X}} 
\toprule
       & $l_2$ & $l_3$ & $l_4$ & $\mathrm{EE}_x$ & $\mathrm{EE}_y$ \\
\midrule
max. & 0.95 & 2.00 & 3.00 & 2.50 & 1.50 \\
min. & 0.05 & 0.05 & 0.05 & -0.50 & -1.50 \\
\bottomrule
\end{tabularx}
\end{center}
\end{table}


\subsection{Conditions}
\label{subsec:mechanism-conditions}
Due to the difference in linkage lengths, mechanisms naturally have different mechanical properties that may be used to assess their performance. 
This study aims to create numerous and distinct linkage mechanisms that satisfy the given performance conditions. 
In other words, linkage mechanisms produced given similar conditions must be performance-compliant, yet still different to each other (diversity).
Thus, a means to evaluate the performance and uniqueness of the link sets is required. 
This section explains the kinematic and quasi-static conditions of the four-bar linkage mechanisms we considered in this paper.

\paragraph{Kinematic Condition}
As shown in Fig.~\ref{fig:linkage-workspace}, the maximum distance available in the path a mechanism draws ($d_{max}$) can be derived by numerically calculating the Euclidean distances between the points along the path.

\paragraph{Quasi-static Condition}

In the real world, there are many cases where the quasi-static aspects of a mechanism must also be considered, e.g., delivering a particular load from one position to another in a desired path or workspace.
In order to determine the maximum payload of a mechanism for a task, we calculate the torque per force ratio between the input link (crank) and the end-effector along a path.

The forces acting in a workspace is determined by the transmission of input torque, and their relationships can be known using the work conservation law.
Assuming the energy is conserved within the mechanism, the relationship between the input and the output work is defined as $|W_{in}| = |W_{out}|$, where $W_{in}$ is the input work at the crank and $W_{out}$ is the output work at the end-effector.
As $W = F \cdot S $ where $F$ is the force and $S$ is the displacement, this equation can be converted to Eq.~(\ref{eq:work-conservation}).

\begin{equation}
    \label{eq:work-conservation}
    |F_{in} \cdot \delta S_{in}| = |F_{out} \cdot \delta S_{out}|
\end{equation}
As $F\cdot S = \tau \cdot \theta$ where $\tau$ is torque and $\theta$ is angular displacement, it can further be derived as Eq.~(\ref{eq:force-to-torque}).

\begin{equation}
    \label{eq:force-to-torque}
    |\tau_{in} \cdot \delta \theta_{in}| = |F_{out} \cdot \delta S_{out}|
\end{equation}
where $\delta \theta_{in}$ is the incremental angular displacement at the crank and $\delta S_{out}$ is the incremental displacement at the end-effector.
Since we defined $\eta$ as the torque per force ratio, $\eta$ can be expressed as Eq.~(\ref{eq:eta}).

\begin{equation}
    \label{eq:eta}
    \eta = \left| \frac{\delta S_{out}}{\delta \theta_{in}} \right|
\end{equation}

Equation~\ref{eq:eta} can now be used to calculate the $\eta$ at the end-effector where a payload will be attached to.
Figure~\ref{fig:vis-eta-min}~(a) shows that $\eta$ of a mechanism is determined by the incremental angular displacement of the crank ($\delta \theta$) and the incremental displacement of the end-effector ($\delta S_{out}$).
As this study solely considers a fully-revolving crank-rocker mechanism, $\eta$ along its full path can be calculated as shown in Fig.~\ref{fig:vis-eta-min}~(b). The workspace of the crank-rocker mechanism has a closed curve, so dividing it based on the maximum linear distance it can move in the workspace will give two paths. Each of these two paths will have a different $\eta_{min}$ for each path. If we divide the path by two at the maximum distance points (red points in Fig.~\ref{fig:vis-eta-min} (b)), two different $\eta_{min}$ can be determined at each path (blue in Fig.~\ref{fig:vis-eta-min}~(b)).
Although a crank-rocker can fully revolve, it can also move only in one of these two paths.
In other words, one of the two paths that gives the higher $\eta_{min}$ can be selected to maximize the payload.
By using this measure, we can determine the minimum torque requirement of a motor to lift desirable payload, and vice versa.

Figure~\ref{fig:vis-sampled-linkage} shows that there are many different samples that satisfy the same kinematic condition, despite their different linkage lengths and paths. It is important to note that there are many different mechanism samples, even if it has very similar conditions. Therefore, a variety of samples can be generated when the linkage is generated along the selected $d_{max}$.

\subsection{Data Generation for Model Training}
\label{subsec:mechanism-generation}

\begin{figure}[tb]
\centering
\includegraphics[width=0.45\textwidth]{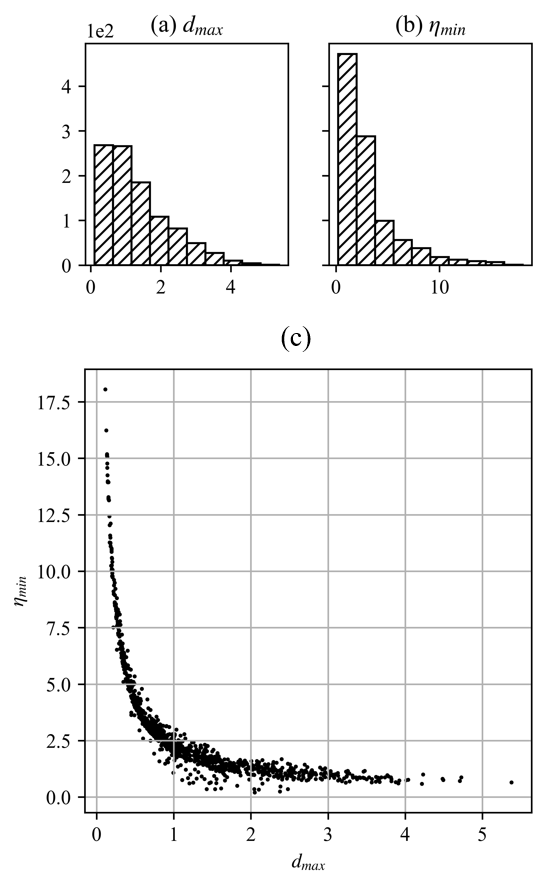}
\caption{(a), (b) Histograms and (c) a scatter plot of $d_{max}$ and $\eta_{min}$ of $X_{train}$}
\label{fig:hist-scatter-real-samples-merged}
\end{figure}

\begin{figure}[bth]
\centering
\includegraphics[width=0.45\textwidth]{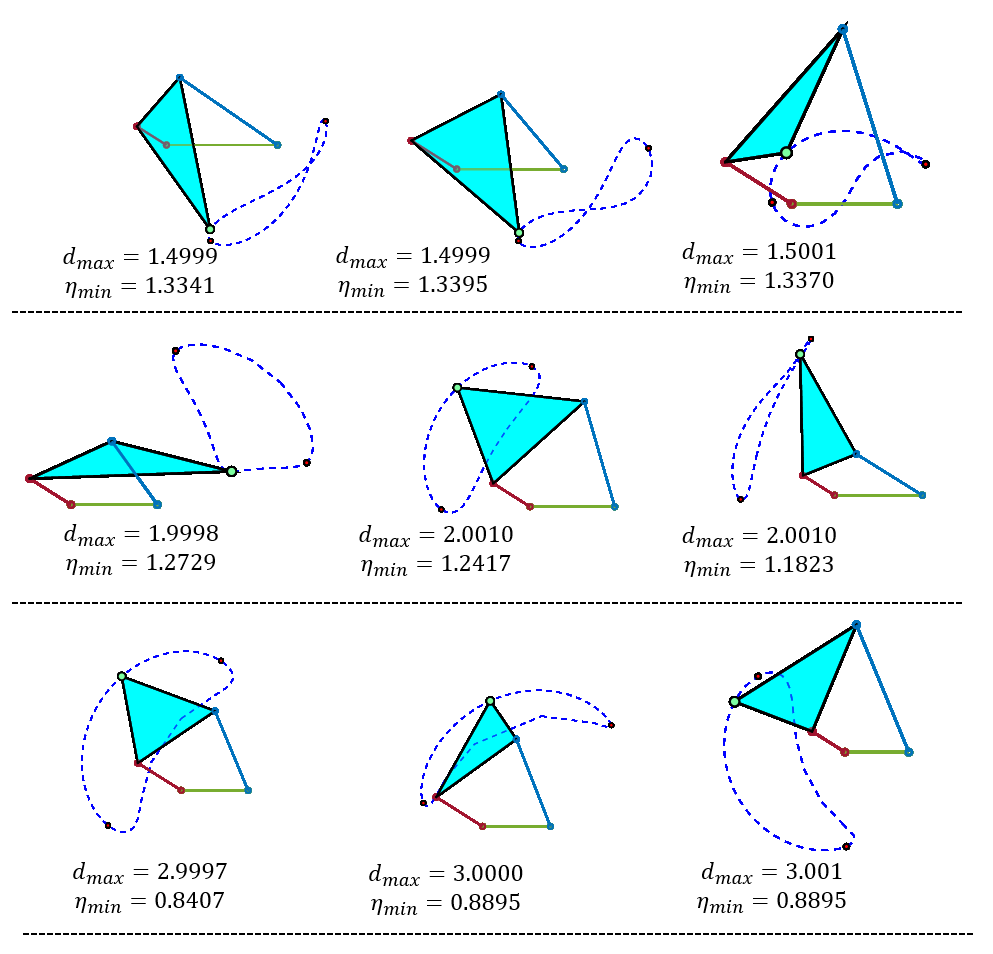}
\caption{Visualization of few mechanisms with similar conditions generated for the model training}
\label{fig:vis-sampled-linkage}
\end{figure}

The dataset of the linkage mechanisms used for the training of a generative model and synthesis was produced using Latin Hypercube Sampling (LHS)~\parencite{mckay1979lhs} within the arbitrary linkage length ranges defined in Table~\ref{tab:length-range}.
The goal of using LHS is to ensure that the samples are representative of the distribution across all ranges of linkage lengths, while minimizing any correlations between them.
Any generated samples that do not satisfy the crank-rocker conditions given in Eq.~(\ref{eq:linkage-conds}) were removed. 100,000 valid samples were generated in total. We denote this dataset for the model training as $X_{train}$.
The generation of $X_{train}$ was done in \MATLAB. Figure~\ref{fig:hist-scatter-real-samples-merged} shows the distribution of $d_{max}$ and $\eta_{min}$ calculated from $X_{train}$.
As shown, $d_{max}$ and $\eta_{min}$ naturally have a non-linear relationship that was not apparent before the generation.
A few of the generated mechanisms with similar conditions are visualized in Fig.~\ref{fig:vis-sampled-linkage} to show their diversity in linkage lengths.

\section{Synthesis with Deep Generative Model}
\label{sec:synthesis-with-generative-model}

\begin{figure*}[tb]
    \centering
    \includegraphics[width=\textwidth]{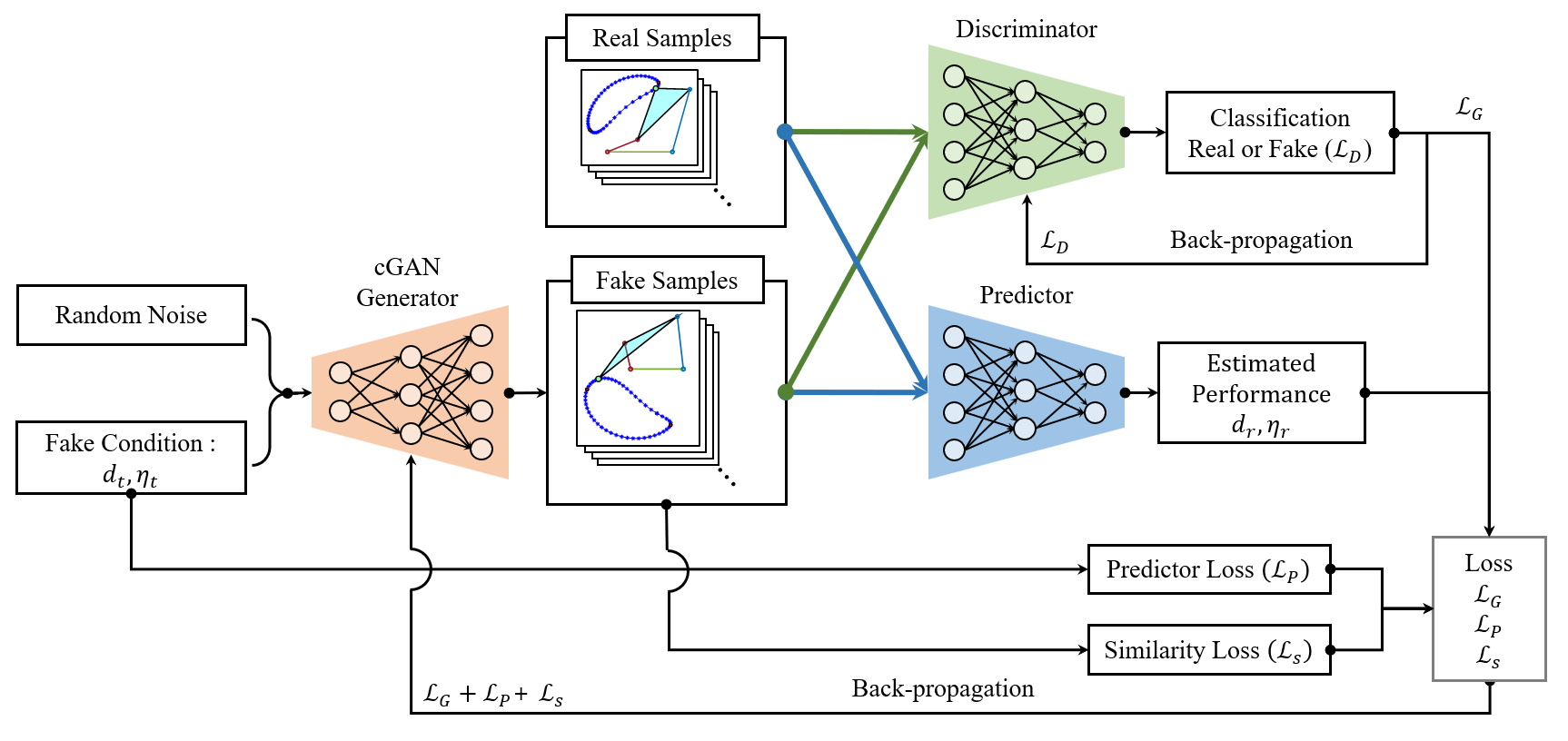}
    \caption{The training procedure (loop from left to right) of the modified cGAN model}
    \label{fig:cgan-training}
\end{figure*}

cGAN~\parencite{goodfellow2020gan,mirza2014cgan}, a popular deep learning-based generative model, was used to synthesize mechanism samples for given conditions.
cGAN has a strength compared to other deep generative models in that it can generate multiple realistic samples that satisfy a given condition. 
This section explains the model training with $X_{train}$, the mechanism synthesis, and the evaluation of the trained model.

\subsection{Conditional Generative Adversarial Network}
\label{subsec:structure-and-training-of-cGAN}

\paragraph{Training}
 
The training procedure of the cGAN model is shown in Fig.~\ref{fig:cgan-training}. 
The generator ($G$) generates synthetic design samples for given random conditions, where the conditions are $d_{max}$ and $\eta_{min}$. 
The random noise input is to ensure that random samples are generated. 
The discriminator ($D$) then classifies unlabeled real and synthesized (fake) samples on whether they are real or fake. 
This process is looped, and $G$ and $D$ are adversarially trained, which enables the generator to synthesize realistic samples. 
The samples consist of link lengths (\(l_2\), \(l_3\), \(l_4\), \(\mathrm{EE}_x\) and \(\mathrm{EE}_y\)) and the conditions. 
The difference between real and fake samples is that the real conditions are the calculated ground truths for the given link lengths, whereas link lengths are estimates that depend on randomly generated conditions for fake samples.

\paragraph{Structure}

$G$ is a multi-layer perceptron (MLP) with five fully-connected layers with 20 neurons each.
Note that the number of layers and neurons are hyperparameters, and the best ones that perform reasonable inverse design were found by trial-and-error.
A Rectifier Linear Unit (ReLU)~\parencite{glorot2011relu} activation function was used between the layers. 
$D$ is identical to $G$, except that it has an additional sigmoid activation function at the end to enforce the output between 0 and 1 to calculate classification losses.
A Binary Cross-Entropy (BCE) loss was used for $D$ ($\Loss_D$).
The batch size was set to 100.

\paragraph{Modifications on cGAN}
The baseline cGAN was first tested to synthesize mechanism samples.
However, the model could not generate good samples due to the infamous mode collapse, creating near identical samples for given conditions regardless of hyperparameters.
Since the aim is to generate diverse yet accurate samples, additional methods were applied to the baseline cGAN model.

A predictor ($P$) was used to predict the estimated actual $d_{max}$ and $\eta_{min}$ ($d_r$ and $\eta_r$) for given samples.
This was used to minimize the gap between the $d_{max}$ and $\eta_{min}$ given as target conditions ($d_t$ and $\eta_t$) to $G$ and $d_r$ and $\eta_r$ of the samples that $G$ synthesized.
The gap was measured with mean squared error (MSE) loss ($\Loss_P$) of a batch as shown in Eq.~(\ref{eq:loss-l-P}), which was used as a part of loss functions for $G$.
$P$ is also an MLP identical to $G$, except that its inputs are the linkage mechanism samples, and the outputs are the estimated conditions.
It is worth noting that $P$ can be simply replaced with the script that was used to calculate the conditions of $X_{train}$ in Section~\ref{subsec:mechanism-conditions}.
However, this will make the training of cGAN too slow as the script takes a long time to calculate.
Therefore, $P$ was trained using the existing data ($X_{train}$) and the corresponding conditions that have been already calculated.
$P$ was trained prior to the training of cGAN and was fixed with no updates while training the cGAN model.

\begin{equation}
    \label{eq:loss-l-P}
    \Loss_P = \frac{1}{N} \sum^{N}_{i=1} \left[(d_{r,i} - d_{t,i})^2 + (\eta_{r,i} - \eta_{t,i})^2\right]
\end{equation}

Similarity loss ($\Loss_s$) was used to ensure that $G$ generates distinct samples. It is the mean of the nearest Euclidean distances between synthesized samples shown in Eq.~(\ref{eq:loss-s}) to maximize the diversity of samples.

\begin{equation}
    \label{eq:loss-s}
    \Loss_s = - \frac{1}{N} \sum^{N}_{i=1} \min{D \left( \mathrm{x}_i, \mathrm{x}_j \right)}
\end{equation}
where $D$ is the Euclidean distance, and $(\mathrm{x}_i, \mathrm{x}_j)$ are a pair of synthesized samples.

Therefore, the loss function used to train $G$ is as shown in Eq.~(\ref{eq:loss-G}), where $w_P$ and $w_s$ are the weights for $\Loss_P$ and $\Loss_s$, respectively, and were adjusted accordingly to stabilize the training.

\begin{equation}
    \label{eq:loss-G}
    \Loss_{GPs} = \Loss_G + w_P \Loss_P + w_S \Loss_s
\end{equation}

After the training, the samples can be generated given the values of $d_t$ and $\eta_t$, and the respective $d_r$ and $\eta_r$ can be calculated.

\begin{figure}[tb]
    \centering
    \includegraphics[width=0.5\textwidth]{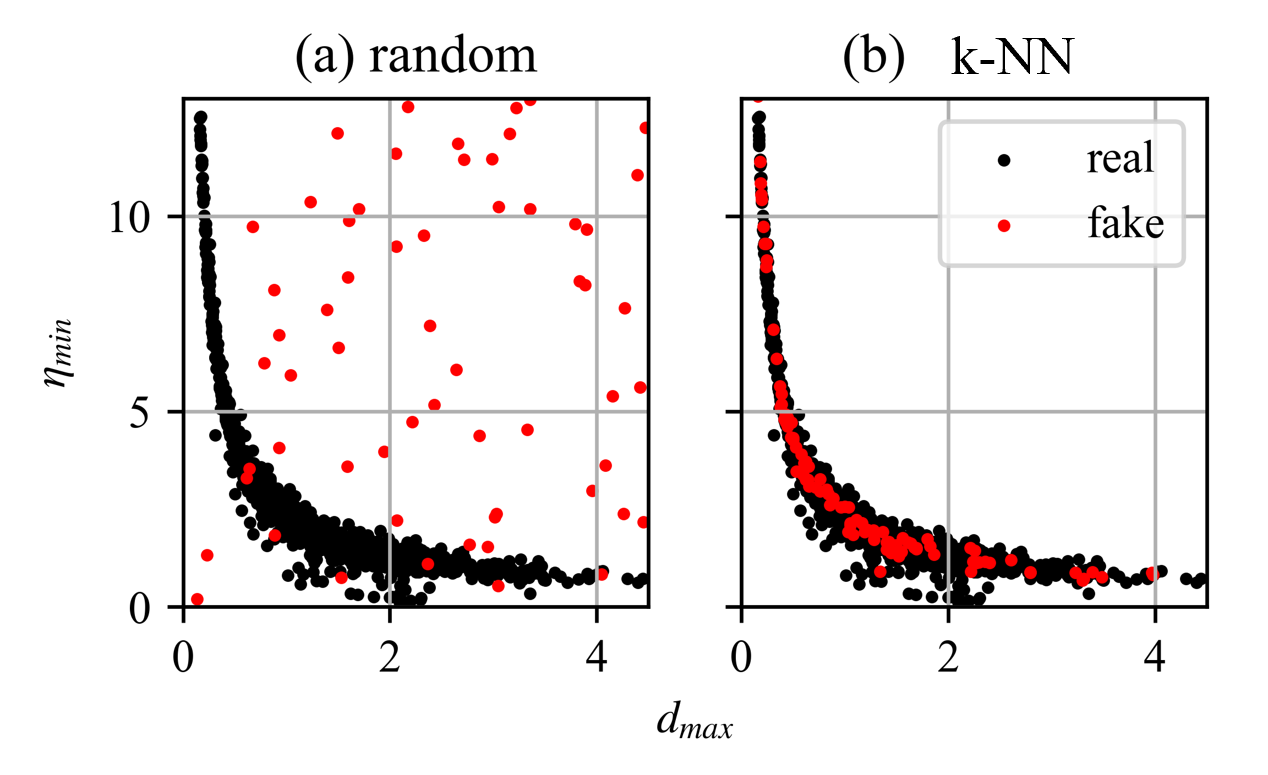}
    \caption{A comparison of the sampling strategies for fake conditions: (a) random and (b) k-NN}
    \label{fig:scatter-fake-condition-sampling}
\end{figure}

It is important to sample realistic fake conditions for $G$, as $D$ could simply discriminate fake samples based on their unrealistic fake conditions.
A random sampling of fake conditions was initially conducted to sample fake conditions for the training of the generative model.
However, as shown in Fig.~\ref{fig:hist-scatter-real-samples-merged}~(c), the real conditions have Pareto-like distribution and make random sampling unreasonable.
This necessitated a more sophisticated sampling strategy.

Synthetic Minority Oversampling Technique (SMOTE)~\parencite{chawla2002smote} is widely used to mitigate imbalanced classification problems by oversampling based on real samples. 
SMOTE works by creating synthetic examples of the minority class by interpolating between existing examples, known as the k-Nearest Neighbors algorithm (k-NN). We used a similar technique based on k-NN to sample fake conditions for the modified cGAN model. This allows for more realistic sampling which can be used effectively in the proposed framework.  As shown in Fig. \ref{fig:scatter-fake-condition-sampling}, the fake conditions generated by LHS much less realistic. Therefore, generating samples with LHS is likely to be a mechanically non-existent solution.

\subsection{Evaluation}
\label{subsec:cGAN-evaluation}

\begin{figure}[tb]
    \centering
    \includegraphics[width=0.45\textwidth]{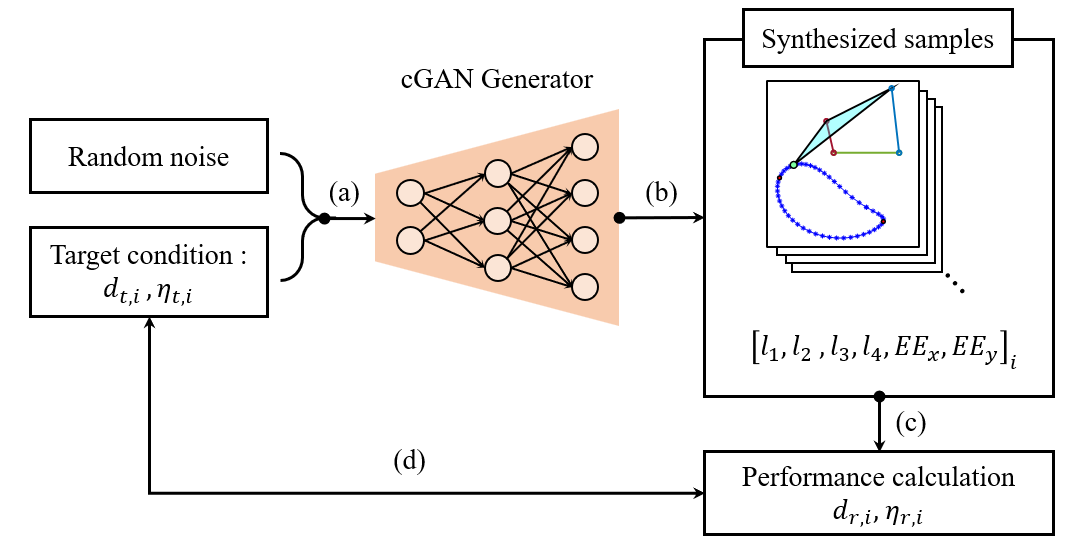}
    \caption{The evaluation procedure of generator ($G$)}
    \label{fig:cGAN-evaluation-procedure}
\end{figure}

Figure~\ref{fig:cGAN-evaluation-procedure} shows the evaluation procedure of the trained model.
The procedure is as follows:

\textbf{(a)}~The number of $i$ fake conditions are sampled. 
Equation~\ref{eq:eval-target-multiple} states a set of fake (target) conditions $\left[ \mathcal{D}_{t}, \mathcal{H}_{t} \right]$.
\begin{equation}
    \left[ \mathcal{D}_{t}, \mathcal{H}_{t} \right] = \left[ \{d_{t,1}, d_{t,2}, ..., d_{t,i}\}, \{\eta_{t,1}, \eta_{t,2}, ..., \eta_{t,i}\} \right]
    \label{eq:eval-target-multiple}
\end{equation}

\textbf{(b)}~The linkage mechanism length samples are synthesized with $G$ with $\left[ \mathcal{D}_{t}, \mathcal{H}_{t} \right]$.
We denote this dataset $X_{syn}$.

\textbf{(c)}~$d_{r}$ and $\eta_r$ values of $X_{syn}$ are calculated.
Equation~\ref{eq:eval-real-multiple} states a set of actual conditions $\left[ \mathcal{D}_{r}, \mathcal{H}_{r} \right]$ calculated from $X_{syn}$.
\begin{equation}
    \left[ \mathcal{D}_{r}, \mathcal{H}_{r} \right] = \left[ \{d_{r,1}, d_{r,2}, ..., d_{r,i}\}, \{\eta_{r,1}, \eta_{r,2}, ..., \eta_{r,i}\} \right]
    \label{eq:eval-real-multiple}
\end{equation}

\textbf{(d)}~Finally, the model is evaluated by calculating errors between $\left[ \mathcal{D}_{t}, \mathcal{H}_{t} \right]$ and $\left[ \mathcal{D}_{r}, \mathcal{H}_{r} \right]$.

The trained model was tested with two cases.
Firstly, 100,000 fake conditions were sampled with k-NN  to test the model performance in generating mechanism samples that satisfy the given target conditions.
Secondly, 100 samples ($i=100$) with a single (same) condition were used to test the ability of the model to produce diverse mechanism samples.

\paragraph{Multiple Conditions}

\begin{table}[h!]
\caption{The performance of the trained model for multiple conditions}
\centering
\label{tab:cGAN-error-multiple}
\begin{tabularx}{\linewidth}{c|*{2}{>{\centering\arraybackslash}X}} 
\toprule
 & $\left[ \mathcal{D}_{t}, \mathcal{D}_{r} \right]$ & $\left[ \mathcal{H}_{t}, \mathcal{H}_{r} \right]$ \\
\midrule
RMSE & 0.147 & 0.894 \\
MAE & 0.110 & 0.398 \\
$R^2$ & 0.958 & 0.850 \\
\bottomrule
\end{tabularx}
\end{table}


Root mean squared error (RMSE), mean absolute error (MAE), and $R^2$ scores were used to evaluate the performance of the trained cGAN model in Table~\ref{tab:cGAN-error-multiple}.
As shown, $G$ was able to generate synthetic linkage mechanisms that satisfy both of the target conditions with marginal errors.

\begin{figure}[tb]
    \centering
    \includegraphics[width=0.45\textwidth]{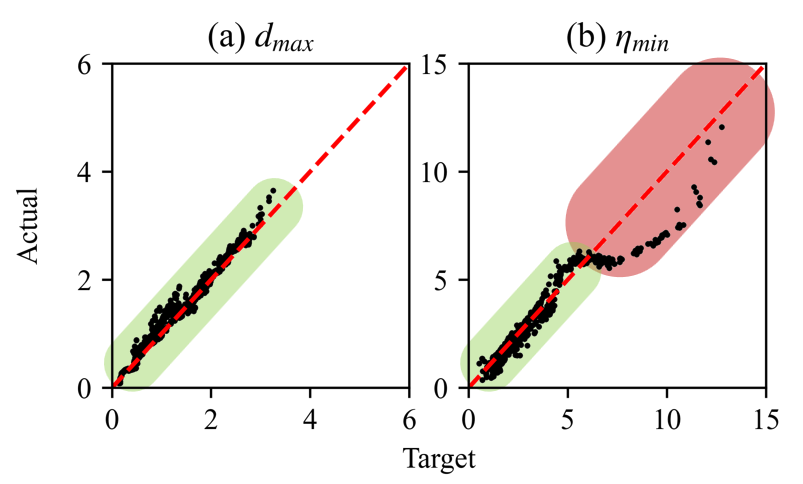}
    \caption{The scatter plots of (a) $d_t$ vs $d_r$ and (b) $\eta_t$ vs $\eta_r$. The dashed line in red is $y=x$.}
    \label{fig:cGAN-scatter}
\end{figure}

Figure~\ref{fig:cGAN-scatter} shows the scatter plots between the target and the actual conditions.
$G$ was mostly able to produce the samples that satisfy the target conditions.
However, it struggled to match $\eta_t$ when $\eta_t > 6$.
We suspect this is due to the lack of data in this region; as shown in Fig.~\ref{fig:hist-scatter-real-samples-merged}~(b), $X_{train}$ naturally has much more data where $\eta_{min} < 6$, even though $X_{train}$ was generated with random linkage lengths.
This means that $G$ would have been trained to synthesize better where $d_{max}$ and $\eta_{min}$ are mostly concentrated, as the fake conditions were generated based on the distribution of the conditions of $X_{train}$.
Moreover, $d_{max}$ gets extremely low as $\eta_{min}$ increases, and vice versa, which may not be useful for real-world use.
Therefore, one may want a mechanism that reasonably satisfies both of the requirements, in which proposed model provides.
Nevertheless, a more sophisticated sampling technique for fake conditions or a variable weighting of samples while training $G$ might mitigate this issue.

\paragraph{Single Condition}

\begin{table}[tb]
\caption{The performance of the trained model for the single condition}
\centering
\label{tab:cGAN-error-single}
\begin{tabularx}{\linewidth}{c|*{2}{>{\centering\arraybackslash}X}} 
\toprule
 & $\left[ \mathcal{D}_{t}, \mathcal{D}_{r} \right]$ & $\left[ \mathcal{H}_{t}, \mathcal{H}_{r} \right]$ \\
\midrule
RMSE & 0.265 & 0.295 \\
MAE & 0.174 & 0.262 \\
\bottomrule
\end{tabularx}
\end{table}

\begin{table}[tb]
\caption{The standard deviation of the synthesized sample lengths for the single condition}
\centering
\label{tab:std}
\begin{tabularx}{\linewidth}{*{5}{>{\centering\arraybackslash}X}} 
\toprule
$l_2$ & $l_3$ & $l_4$ & $\mathrm{EE}_x$ & $\mathrm{EE}_y$ \\
\midrule
0.041 & 0.036 & 0.138 & 0.644 & 0.168 \\
\bottomrule
\end{tabularx}
\end{table}

The condition set was fixed to $d_t = 1.0, \eta_t = 2.0$ to evaluate the model performance in generating diverse samples.
In other words, $\left[ \mathcal{D}_{t}, \mathcal{H}_{t} \right]$ now have the same values regardless of $i$.
Table~\ref{tab:cGAN-error-single} shows the error metrics of the model given the same condition.
Note that the $R^2$ metric is not used as the conditions are the same in this case.
The standard deviation of the synthesized sample lengths is shown in Table~\ref{tab:std}.
From these metrics, it can be seen that the model is able to synthesize diverse mechanism samples while satisfying the target conditions with marginal errors.

\section{Ablation Studies}\label{sec:ablation-studies}
We have proposed few modifications on the baseline cGAN to facilitate the generation of linkage mechanisms.
In Sec.~\ref{ssec:modifications-on-cgan}, we evaluate the effectiveness of these modifications by performing a quantitative performance analysis of cGAN with and without these modifications.
Moreover, in Sec.~\ref{ssec:model-perf-comp}, we also compare our model with other well-established optimization and generative models to assess our contributions.

\subsection{Modifications on cGAN}
\label{ssec:modifications-on-cgan}
This study proposed few modifications on the baseline cGAN in Section~\ref{subsec:structure-and-training-of-cGAN} to improve the synthesis performance for four-bar mechanism synthesis.
In this section, we compare the performance of the cGAN with and without these modifications and assess the effectiveness of the suggested modifications.
Table \ref{tab:ablation-study-models} shows the four models built for the ablation study of the proposed cGAN. The predictor loss ($\Loss_P$), similarity loss ($\Loss_s$), and k-NN are the three key elements we introduced to improve the performance of cGAN for mechanism synthesis.
k-NN, however, was not included in this ablation study as it was impossible to train a model well without it due to the unique relationship between conditions as shown in Fig.~\ref{fig:hist-scatter-real-samples-merged}~(c).
In other words, a more advanced condition sampling strategy than a uniform or Gaussian sampling must be used to train a conditional generative model well for mechanism synthesis due to such nature.

\begin{table}[tb]
  \caption{The four cGAN models built for the ablation study on the added elements of cGAN}
  \centering
  \label{tab:ablation-study-models}
  \begin{tabularx}{\linewidth}{c|*{2}{>{\centering\arraybackslash}X}}
    \toprule
    \textbf{Model} & \textbf{Predictor ($\Loss_P$)} & \textbf{Similarity ($\Loss_s$)}\\
    \midrule
    A & X & X\\
    B & O & X\\
    C & X & O\\
    D & O & O\\
    \bottomrule
  \end{tabularx}
\end{table}

For the best results, we performed hyperparameter optimization for the model training.
The hyperparameters to be optimized are the learning rates of $G$ and $D$, and the weights for the predictor loss ($w_P$) and similarity loss ($w_s$).
$[0.1, 0.01, 0.001, 0.0001]$ were used as a grid for all the hyperparameters.
The set of hyperparameters with the lowest $\Loss_P$ was used for the evaluation of each model.
It is not feasible to choose one with the lowest $\Loss_s$ as it is always achieved by making the linkage lengths very long, showing very high $\Loss_P$.

\begin{figure}[tb]
\centering
\includegraphics[width=0.45\textwidth]{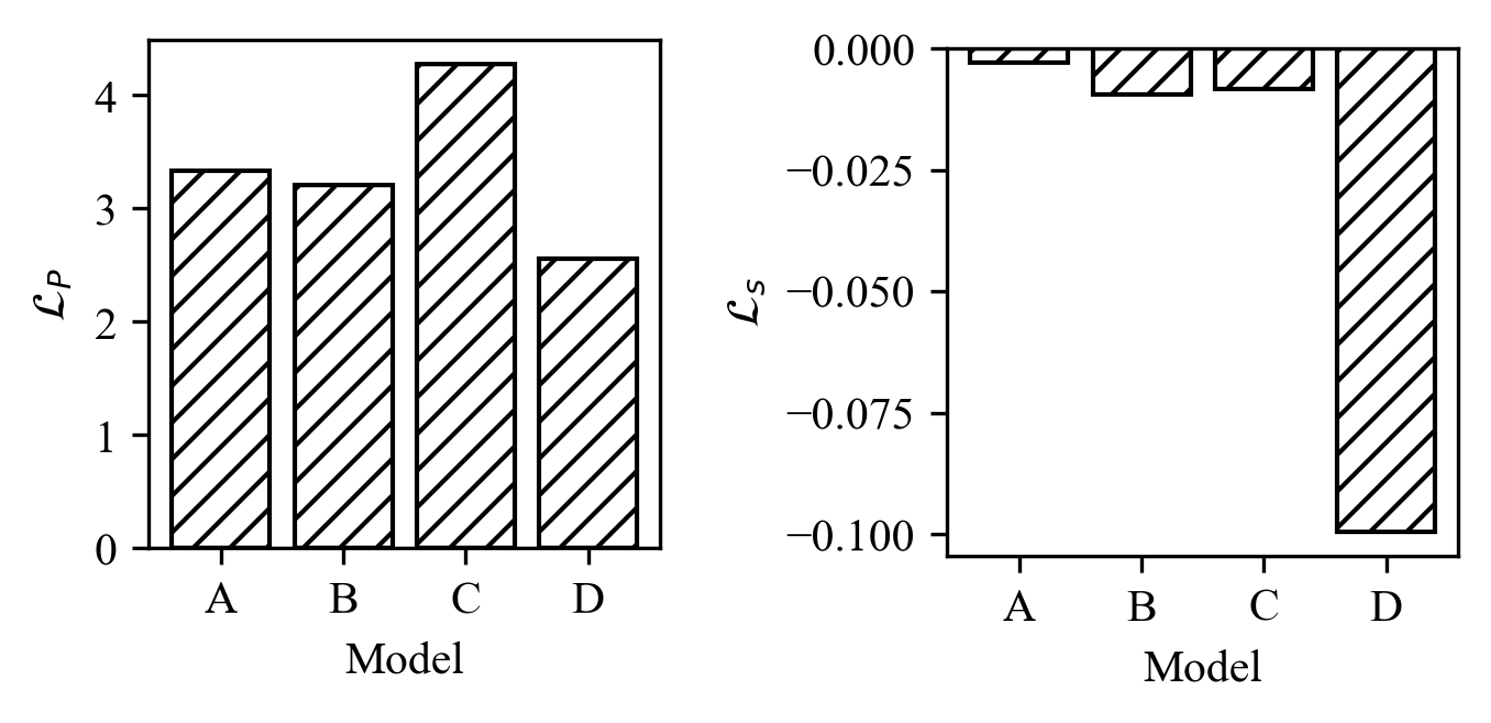}
\caption{$\Loss_P$ and $\Loss_s$ of the ablation models. The lower the better.}
\label{fig:ablation-loss}
\end{figure}

Figure~\ref{fig:ablation-loss} shows the comparison of $\Loss_P$ and $\Loss_s$ of the trained ablation models. 
Compared to the baseline cGAN model (Model A), it is shown that the addition of $P$ (Model B) improves $\Loss_P$, while the sole introduction of $\Loss_s$ (Model C) slightly improves $\Loss_s$ while sacrificing $\Loss_P$.
The improvements were more noticeable when both $\Loss_P$ and $\Loss_s$ were applied (Model D), showing much better results than Model A.


\begin{table*}[tb]
\begin{center}
\caption{The synthesis performance comparison of the proposed cGAN, traditional cVAE and NSGA-II for the target conditions}
\label{tab:model-comp}

\begin{tabularx}{\linewidth}{X|XX|XXX|XXX|XXXX} 
\toprule
     & & & \multicolumn{3}{c|}{cGAN} &\multicolumn{3}{c|}{cVAE} &\multicolumn{4}{c}{NSGA-II} \\ 
     & $d_t$ & $\eta_t$ & RMSE & MAE & $\Loss_s$ & RMSE & MAE & $\Loss_s$ & RMSE & MAE & $\Loss_s$ & invalid (\%) \\
\midrule
    (a) & 0.4 & 5.0 & 0.340 & 0.213 & \textbf{-0.114} & 0.708 & 0.455 & -0.005 & \textbf{0.059} & \textbf{0.042} & 0.000 & 4 \\
    (b) & 1.0 & 2.0 & 0.295 & 0.235 & \textbf{-0.127} & 0.555 & 0.468 & -0.013 & \textbf{0.059} & \textbf{0.059} & -0.001 & 2 \\
    (c) & 2.0 & 1.5 & \textbf{0.361} & \textbf{0.273} & \textbf{-0.132} & 0.826 & 0.804 & -0.011 & 0.480 & 0.351 & 0.000 & 0 \\
    (d) & 2.5 & 1.0 & 0.337 & 0.286 & \textbf{-0.125} & 1.091 & 1.064 & -0.011 & \textbf{0.030} & \textbf{0.016} & -0.093 & 87 \\
\midrule
    Avg. & & & 0.333 & 0.252 & \textbf{-0.125} & 0.795 & 0.698 & -0.010 & \textbf{0.278} & \textbf{0.147} & -0.003 & 23 \\
\bottomrule
\end{tabularx}

\end{center}
\end{table*}

\begin{figure*}[t!]
    \centering
    \includegraphics[width=\textwidth]{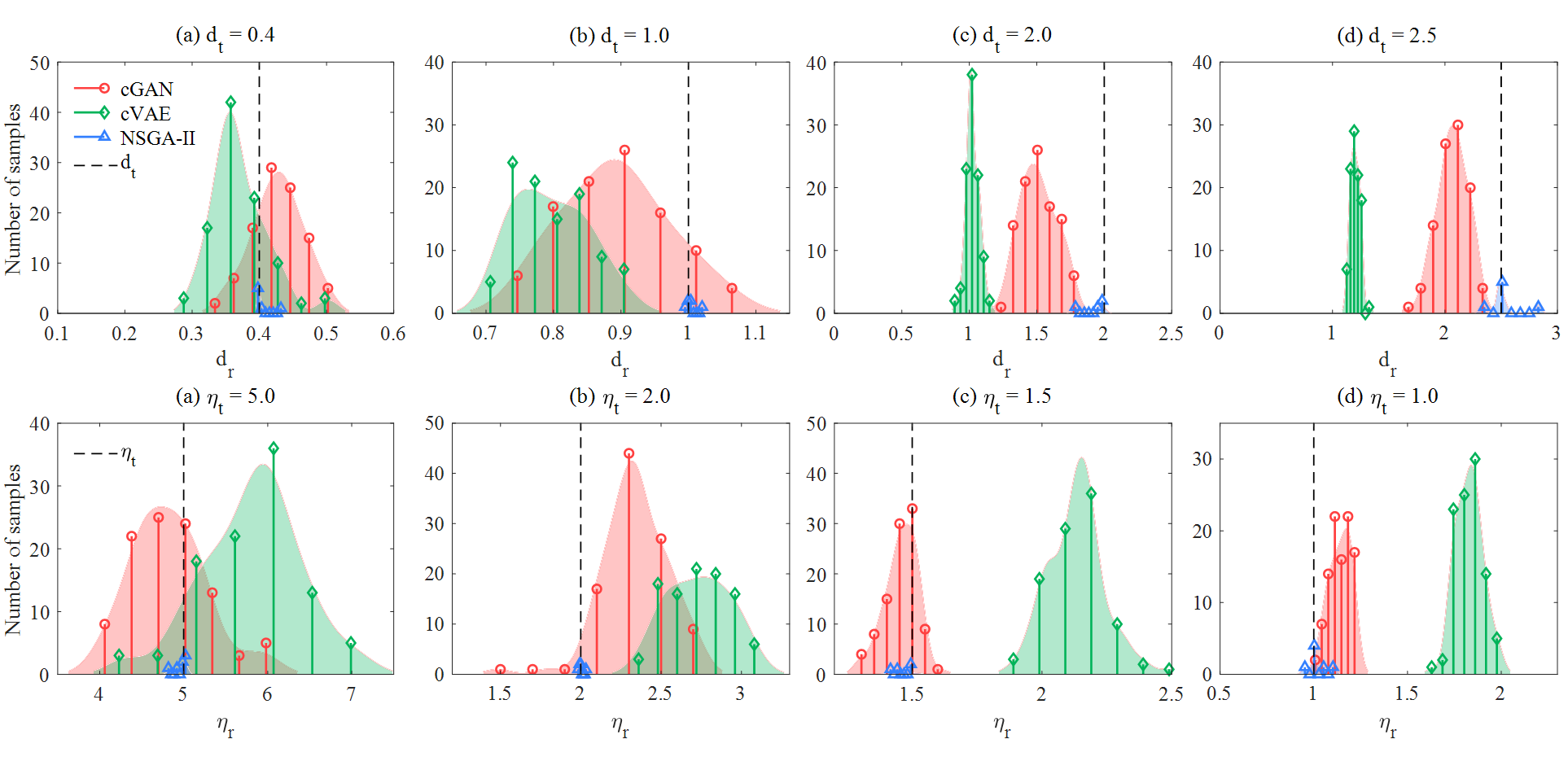}
    \caption{The distribution of $d_r, \eta_r$ of the generated mechanism samples for cGAN, VAE, and NSGA-II for the given target conditions. In NSGA-II, the duplicate samples among the generated samples were merged into one for visualization due to the model generating too many duplicates.}
    \label{fig:hist-model-comp}
\end{figure*}

\begin{sidewaysfigure*}
    \centering
    \includegraphics[width=1\textwidth]{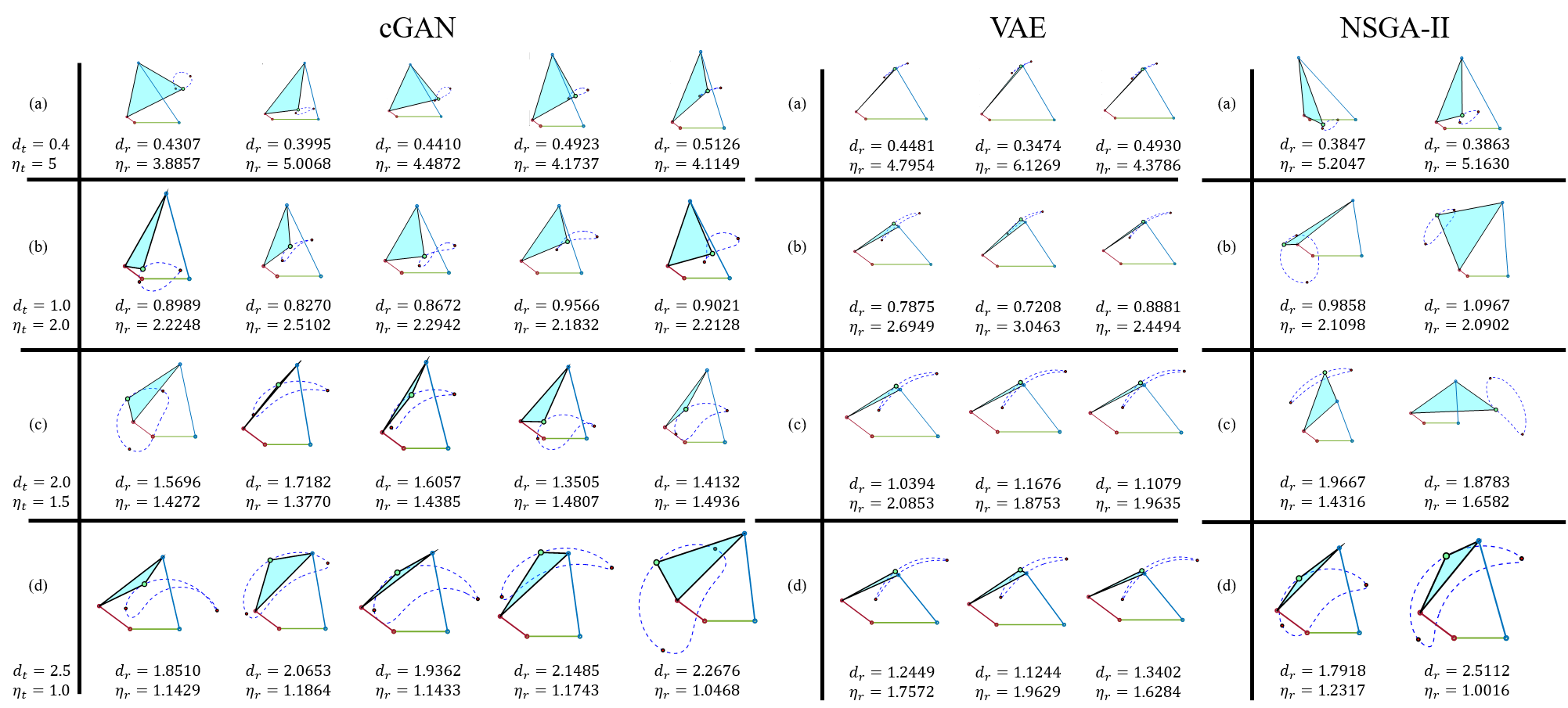}
    \caption{A comparison of the generated samples from the proposed cGAN, NSGA-II, and cVAE for the given target conditions}
    \label{fig:model-comp}
\end{sidewaysfigure*}

\subsection{Model Performance Comparison}
\label{ssec:model-perf-comp}

 To assess our contributions for the generation of target-compliant linkage mechanisms, we compare the performance of the proposed model for mechanism synthesis to other well-established models.
 non-dominated sorting genetic algorithm-II (NSGA-II) \parencite{deb2022nsgaii}, a popular evolutionary optimization algorithm and conditional VAE (cVAE), a deep learning-based generative model similar to cGAN which is often used for mechanism synthesis \parencite{deshpande2019VAE,deshpande2020convVAE}, were chosen for comparison.
 It is important to note that the approach of \cite{deshpande2019VAE} for mechanism synthesis using VAE is different to our approach, thus this should not be considered as a direct performance comparison of our method to theirs.

Although NSGA-II is a potent algorithm capable of generating multiple optimized samples, typical optimization-based algorithms are not designed to generate 100,000 samples in short time.
In other words, it was not possible to generate 100,000 samples with NSGA-II within a reasonable timeframe due to the complex nature of the algorithm and the computational constraints.

Thus, the accuracy and diversity of the selected models for comparison were calculated with 100 samples each for four different target conditions.

Predictor ($P$), the same surrogate model used to train the proposed cGAN, was also used for NSGA-II to evaluate the samples during generation. The quantitative comparison of their performance is shown in Table~\ref{tab:model-comp}.
The proposed cGAN model shows much lower diversity loss ($\Loss_s$) than the other models, meaning that it can generate much more diverse samples, giving more choices to select a suitable mechanism design for one's needs.
Although NSGA-II has shown slightly higher accuracy on average, it is important to note that the model generated many invalid samples, even with much lower diversity of the samples.
No invalid samples were produced with cGAN or cVAE. cVAE was not able to produce more accurate or diverse samples than cGAN or NSGA-II.

Figure~\ref{fig:hist-model-comp} shows the distribution of the resulting $d_r$ and $\eta_r$ of the generated samples.
The generated samples (population) of NSGA-II had a lot of duplicate samples, where less than 10\% of the generated samples were not duplicates, reducing the design choices further.
As shown in the figure, NSGA-II tries to optimize samples to the target as much as possible, whereas cGAN has more diverse distribution that covers larger design space at the cost of slightly lower accuracy.
Figure~\ref{fig:model-comp} is the visualization of few mechanisms generated by the models for qualitative comparison.




\section{Conclusion}\label{sec:conclusion}
In this paper, we proposed a modified cGAN model with a predictor and diversity loss for accurate and diverse mechanism synthesis. This model is capable of generating crank-rocker mechanisms that satisfy given target conditions. The target conditions were selected as kinematic ($d_{max}$) and quasi-static ($\eta_{min}$) conditions to generate mechanisms with diverse workspaces while considering payload that a mechanism should transfer.
The initial 100,000 mechanism samples were first sampled for model training. Although the range of linkage lengths for sampling is unrestricted in theory, for the sake of model testing, the frame length ($l_1$) was fixed to 1.0, so the link lengths can be considered as a length ratio to the frame. The other link lengths were sampled using LHS within an arbitrary range based on the experience. After training the cGAN model with this dataset and testing using multiple different target conditions, it was found that the model is capable of generating diverse mechanism samples efficiently while satisfying the given conditions with high accuracy.

In addition, a comparative study was performed among the proposed cGAN, a traditional cVAE, and NSGA-II. 100 samples were generated for four different target conditions. It was shown that the proposed cGAN model is capable of generating much more diverse samples than the other models with similar levels of accuracy.

\paragraph{Limitations}
Several limitations were found in the mechanism synthesis method proposed in this study. First, the linkage system used in this study assumes a simple conservation system with ignored mass and inertia. Therefore, in order to consider the quasi-static condition for use in real applications, several additional engineering performances must be considered through simulation obtained by a 3D CAD model. Second, it was confirmed that our proposed cGAN model had excellent performance, especially in diversity when compared with cVAE and NSGA-II. However, it still did not have as high accuracy as NSGA-II. Nevertheless, we believe that this issue can be sufficiently mitigated by introducing additional design conditions such as safety factor in the detailed design stage for real-life applications.

Also, since the use of cGAN with minor modifications for mechanism synthesis was proved viable in this paper, other cGAN variants specialized for engineering inverse design, e.g., PcDGAN \parencite{heyrani2021pcdgan} or Mo-PaDGAN \parencite{chen2021mopadgan}, may show better performance than the traditional cGAN for this task.
Further empirical studies with such models may be conducted to produce more compliant and diverse mechanisms. 

\subsection*{Acknowledgements}
This work was supported by the National Research Foundation of Korea grant (2018R1A5A7025409) and the Ministry of Science and ICT of Korea grant (No.2022-0-00969, No.2022-0-00986).

\subsection*{Replication of Results}
The code is available online\footnote{\href{https://github.com/jihoonkim888/LinkGAN}{https://github.com/jihoonkim888/LinkGAN}}.

\subsection*{Conflict of interest statement}
The authors declared no potential conflicts of interest with respect
to the research, authorship, and/or publication of this article.

\printbibliography
\end{document}